%% file: UMAP2020.tex
\documentclass[sigconf]{acmart}
\usepackage{balance}
\pdfoutput=1

\input{macros}
\usepackage[linesnumbered,ruled,vlined]{algorithm2e}
\usepackage{pst-solides3d}
\usepackage{bbding}
\usepackage{pifont}
\usepackage{wasysym}
\usepackage{amssymb}
\usepackage{float}
\usepackage{subfig}
\usepackage{enumitem}
\usepackage{footmisc}
\usepackage{wrapfig}
\usepackage[skip=0.2pt]{caption}
\usepackage{resizegather}

\copyrightyear{2020}
\acmYear{2020}
\setcopyright{acmcopyright}\acmConference[UMAP '20 Adjunct]{Adjunct Proceedings of the 28th ACM Conference on User Modeling, Adaptation and Personalization}{July 14--17, 2020}{Genoa, Italy}
\acmBooktitle{Adjunct Proceedings of the 28th ACM Conference on User Modeling, Adaptation and Personalization (UMAP '20 Adjunct), July 14--17, 2020, Genoa, Italy}
\acmPrice{15.00}
\acmDOI{10.1145/3386392.3399568}
\acmISBN{978-1-4503-7950-2/20/07}

\makeindex
\begin{document}
\title{Fair Inputs and Fair Outputs: \\The Incompatibility of Fairness in Privacy and Accuracy}

\author{Bashir Rastegarpanah}
\affiliation{\institution{Boston University}}
\email{bashir@bu.edu}

\author{Mark Crovella}
\affiliation{\institution{Boston University}}
\email{crovella@bu.edu}

\author{Krishna P. Gummadi}
\affiliation{\institution{MPI-SWS}}
\email{gummadi@mpi-sws.org}

\renewcommand{\shortauthors}{Rastegarpanah et al.}
\renewcommand{\shorttitle}{The Incompatibility of Fairness in Privacy and Accuracy}

\begin{abstract}
Fairness concerns about algorithmic decision-making systems have been mainly
focused on the outputs
(e.g., the accuracy of a classifier across individuals or groups).
However, one may additionally be concerned with fairness in the inputs.
In this paper, we propose and formulate two properties regarding the inputs of
(features used by) a classifier.
In particular, we claim that fair privacy (whether individuals are all asked to
reveal the same information) and need-to-know (whether users are only asked for
the minimal information required for the task at hand) are desirable properties
of a decision system.
We explore the interaction between these properties and fairness in the outputs
(fair prediction accuracy).
We show that for an \emph{optimal} classifier these three properties are
in general incompatible, and we explain what common properties of data
make them incompatible. 
Finally we provide an algorithm to verify if the trade-off between the
three properties exists in a given dataset, and use the algorithm
to show that this trade-off is common in real data.
\end{abstract}


%\keywords{fairness, privacy, classification}

\maketitle
\input{Introduction}
\input{RelatedWork}
\input{Formulation-arXiv}
\input{Properties}
\input{Trade-offs-arXiv}

\input{Verification-arXiv}
\input{Conclusion-arXiv}
\input{acks}
\bibliographystyle{ACM-Reference-Format}
\bibliography{UMAP2020.bib}
\appendix
\input{proofs}
\input{nonoptimal}
\clearpage
\input{results}

\end{document}

%% file: macros.tex
\usepackage{flushend}
\usepackage{bbm}
\usepackage{amssymb}

\newtheorem{thm}{Theorem}
\newtheorem{cor}{Corollary}
\newtheorem{prop}{Proposition}

\newcommand{\vecx}{\ensuremath{\mathbf x}}

\newcommand{\vecc}{\ensuremath{\mathbf c}}

\newcommand{\spara}[1]{\smallskip\noindent{\bf{#1}}}

\newtheorem{lemma}{Lemma}

%% file: Introduction.tex
\section{Introduction}\label{sec:intro}
As data-driven decision making systems are increasingly used in modern
society in ways that affect individual lives, concerns have been
raised about their ethical implications.
In particular, recent years have witnessed a fast-growing number of
studies on fairness in the decisions made by such systems, including 
works on developing notions to define, measures to quantify,
and mechanism to ensure \emph{ fair outputs}
(i.e., whether a decision system provides an equitable service to all 
of its users or groups of users).
Despite the natural dependence of decision outcomes on data inputs, fairness
concerns that incorporate the \emph{inputs} of decision system are however
less studied.

Traditionally, ethical concerns about the \emph{inputs to}
(i.e., data used by) decision systems have been the focus of ``privacy''
studies, while ethical concerns about the \emph{outputs from} decision systems
have been the focus of ``fairness'' studies.   However, we observe that privacy
and fairness originate from fundamentally different epistemic arguments.
At a high-level, privacy concerns are rooted in a desire to protect individuals
by limiting or enabling control over the information they reveal to the world.
Fairness concerns, on the other hand, are rooted in a desire for equitable
treatment of individuals (or groups of individuals). As such, privacy and
fairness concerns can independently arise for both the inputs used and the
outputs generated by decision systems.

As a motivating example, consider a decision problem where the goal is to decide
whether an applicant should be offered a loan. The decision for each applicant
is made based on answers that are collected to a number of demographic and
financial status questions.
%\footnote{An answer to each question can be
%thought of as revealing the value of a feature in a feature vector
%associated with that applicant.}.
In settings similar to this example, we recognize certain social concerns
regarding the information that is gathered from each applicant.
In particular, we raise two questions motivated by previous
proposals and legal regulations:

First, considering each applicant individually, we ask
\emph{``what information is necessary for (i.e., what questions are
relevant to) solving the decision problem at hand?''}
In the loan eligibility problem for example, it seems unnecessary to
ask about an applicant's height. Moreover, although it may often be
necessary to ask about education level, an applicant who has an excellent
credit score and a secure job may find a question about his education
level to be unnecessary.
Notice that as revealing each piece of information to the decision
system is associated with a potential loss in privacy, this concern
is related to protecting individuals' privacy.

%\spara{Data Minimization in GDPR}
The above consideration is reflected in 
the EU General Data Protection Regulation (GDPR)~\cite{regulation2016regulation}
as a principle called \emph{data minimization}, which is defined as:
\emph{``Personal data shall be adequate, relevant and limited to what is
necessary in relation to the purposes for which they are processed.''}
%Note however that this definition does not specify how to achieve data minimization
%in a particular decision scenario.

The second ethical question arises when comparing the information used
(i.e., set of questions asked) from different applicants. In particular,
we ask \emph{``how can using different pieces of information from different
applicants amount to discrimination?''}
For instance, a loan applicant may find it unfair that she is asked to answer
a different set of questions comparing to another applicant.

In order to study these questions in a concrete setting,
we consider a classifier and a set of input variables (features).
We assume that the classifier is trained using all the features, and we
study properties of the classifier when it is applied to a test set.
Furthermore, we assume that the classifier is able to classify a given data
point using any subset of the input features.
In other words, for a given classification instance
at the test time, the values of only a subset of input features may be revealed to the
classifier, and the remaining feature values are set to \emph{unknown}\footnote{While we
do not make additional assumptions about the classification algorithm, practical examples
of classification with partially known inputs are using models that can handle different sets
of input features (e.g. the naive Bayes), or using some
imputation procedure to estimate unknown feature values.\label{footnote1}}.

We observe that in such scenarios, one may ask
\emph{``What properties can make the set of data inputs
used for each classification instance more socially desirable?''}
Our first contribution is proposing two properties of classifiers regarding
their inputs to address the above question; namely the
\emph{need-to-know principle} and the \emph{fair privacy principle}.
In the following we introduce each principle and their formal definitions
will be presented in section \ref{sec:properties}.

Notice that we are not concerned with \emph{how} a set of
feature values are selected to be used for classifying each instance,
but we are rather interested in checking whether an arbitrary set of features meets these
properties\footnote{In practice, one needs to specify how the features are selected
(e.g., by using methods that are suggested in~\cite{shim2018joint,maliah2018mdp, trapeznikov2013supervised,yang1998feature}).
However, by studying these properties and their interaction regardless of 
the feature selection procedure, we show inherent trade-offs that
cannot be avoided using any feature selection procedure.\label{footnote2}}.

\spara{The need-to-know principle.} 
This property presents one way of formalizing
\emph{``data minimization''}~\cite{regulation2016regulation} in a classification setting.
We propose a formulation that is based on classification accuracy.
Intuitively, the need-to-know principle requires that the decision system use
only {\it the minimal amount of information} that is necessary for classifying
a data point with a certain accuracy.
This may for example result in restricting the use of irrelevant or
proxy features.

The justification for this principle is rooted in respecting the privacy
rights of individuals to not divulge information about themselves that
is not needed for the task at hand.  Such a consideration is an
important argument for emerging privacy regulations in different
countries that require data aggregators to justify the need to collect
information about individuals~\cite{regulation2016regulation,
reidenberg1994setting, us2011health}.

\spara{The fair privacy principle.}
Intuitively, the fair privacy principle requires that the decision system use the
\emph{same information (i.e., data inputs)} about all individuals when making decisions.
Put differently, the fair privacy principle prohibits a decision system from using more or less or 
different pieces of information about different individuals.
The justification for the fair privacy principle is two-fold:

First, we observe that in many scenarios it is preferable to use the same data inputs
for all individuals since it equalizes the opportunity to get beneficial outcomes.
In the loan eligibility problem for example, if a decision system uses different input
features for two different individuals say, Alice and Bob, Alice might wonder if she
might have been offered a loan had she been asked to provide the same inputs as Bob,
and vice versa.

We do not expect this argument to be desirable in every
situation.  For example, in the case of predicting recidivism rates, it
may seem reasonable to ask for more information from one individual
comparing to the others in order to achieve an accurate
prediction. However, in other domains such as recruiting, it is often
considered best practice for all candidates to be asked the same questions,
i.e., provide the same data inputs.  In fact such considerations have been the main
inspiration for \emph{structured interviews}.

Second, note that one approach to achieve ``equitable treatment of individuals''---
as the basic idea behind fairness--- is equitable protection of individuals against disclosure
of their private data.
The authors in~\cite{ekstrand2018privacy} suggest that a desirable
property for a privacy protection mechanism is to provide its protections
equitably to all its subjects.
From this perspective, In decision scenarios where individuals would prefer to not divulge
their private data and there is cost to revealing such data, it is preferable that all individuals
bear equal cost, and our fair privacy principle guarantees such an equatable share of privacy
costs.

Our running assumption in this paper is that the decision system
(classifier) itself is a privacy adversary. This assumption is consistent with
scenarios such as our loan application example.
Thus we do not consider privacy notions that assume an adversary who is different
from the party that collects and processes personal data
(e.g., differential privacy~\cite{dwork2014algorithmic}).

\spara{The trade-off.} Our second contribution lies in exposing the trade-offs in
simultaneously achieving the proposed fairness and privacy considerations for inputs
as well as previously proposed fairness considerations for outputs.
Specifically, after formalizing our proposed principles of need-to-know and
fair privacy, we show that in general, an \emph{optimal} classifier cannot
simultaneously satisfy both principles and achieve fairness in outputs
(defined as equal prediction accuracy for all individuals). 
We then provide a formal specification of all datasets in which this trade-off exists,
and a practically efficient algorithm to verify whether a given dataset presents
the trade-off.

While each of need-to-know and fair privacy is a desirable property by itself
and it is natural to seek a classifier that satisfies both, we further explain
why achieving these two properties simultaneously is particularly interesting.
Assume a classifier is applied to solve our loan eligibility example. One may 
decide to achieve fair privacy by asking all applicants to provide answers to all
the input features. While this trivial approach will satisfy fair privacy, we
observe that for all applicants whose prediction would not change if using a
subset of feature values, the need-to-know principle is violated\footnote{Another
solution is to use a trivial classifier that does not use any
feature values from  all applicant. Notice that this trivial classifier
violates the adequacy requirement of input data in the ``data minimization'' principle.}.

Therefore, imposing need-to-know constraint can be seen as a way to
eliminate trivial solutions for achieving fair privacy. On the other hand,
using the same subset of input features for all the applicants such that need-to-know is respected will affect the prediction accuracy of those applicants for whom more data inputs are required.
Our incompatibility result in fact formalizes this intuitive argument.

Finally, note that although optimal classifiers are rarely used in practice,
our results pose a new challenge to the design of classifiers that aim at optimality:
``how much one needs to compromise on optimality in order to simultaneously
achieve fairness in the inputs and outputs of a classifier?''

%% file: RelatedWork.tex
\section{Related Work}\label{sec:relwork}
While some recent work has focused on both privacy and fairness
considerations for outputs~\cite{Cummings:2019:CPF:3314183.3323847,
feldman2015certifying, melis2019exploiting, bagdasaryan2019differential,
jagielski2019differentially},
relatively little work (e.g.,~\cite{grgic2018human}) has examined fairness considerations for
inputs.
%and none (to our knowledge) has simultaneously
%accounted for fairness and privacy considerations for inputs.
In this paper we introduce new notions that simultaneously capture both
privacy and fairness properties of inputs in algorithmic decision systems,
and explore their interaction with fairness properties of outputs.
In the following we review related work on different societal aspects of
decision-making systems including privacy and fairness.
  
\spara{Fairness in algorithmic decision-making.}
In recent years several empirical studies have shown how algorithmic decision systems are prone to unfair treatment of their users in
different areas (e.g., online advertisement~\cite{sweeney2013discrimination} and criminal justice~\cite{larson2016we}).
For more examples we point the interested readers to survey papers~\cite{romei2014multidisciplinary, barocas2016big}.
These findings have raised awareness about the importance of fair decision-making systems by regulatory authorities as well~\cite{executive2014big, regulation2016regulation}.

Research on fair classification can be divided into two parts: formulating fairness nations and measures, and developing techniques to improve the fairness of algorithmic systems. Fairness notions can be categorized as those measuring group unfairness and those measuring unfairness at the level of individuals~\cite{speicher2018unified,dwork2012fairness}. Fairness-enhancing techniques in general fall into three categories based on the stage of the classification pipeline that they are employed:
(i) {\it pre-processing}~\cite{kamiran2012data,calmon2017optimized}, (ii) {\it in-processing}~\cite{kamishima2011fairness,agarwal2018reductions,zafar2017fairness}, and (iii) {\it post-processing}~\cite{hardt2016equality, corbett2017algorithmic}.
In this paper we take the accuracy equality~\cite{verma2018fairness} approach to fairness in section~\ref{sec:fair-accuracy}, and we apply it at the individual level as it has been done previously in \cite{rastegarpanah2019fighting,speicher2018unified}.

\spara{Privacy.}
Privacy in information systems is generally understood using two concepts: limitation theory and control theory~\cite{tavani2007philosophical}. Using those theories, several methods have been proposed to protect the privacy of the users in practice such as differential privacy~\cite{dwork2014algorithmic} k-anonymity~\cite{aggarwal2005k} and cryptography~\cite{stinson2005cryptography}.  The definitions of fairness in privacy that we put forward are concerned with which information about a user is used by the classifier, and so relate most closely to limitation theory.

\spara{Need-to-know as a privacy notion.} Achieving \emph{complete privacy} has been the goal of
cryptography approaches such as secure multi-party computation (SMC). However, due to practical constraints
such as computational efficiency and auditing purposes, the alternative goal of acquiring minimum neccessary data
has become important as stated by regulations in different countries~\cite{regulation2016regulation, reidenberg1994setting, us2011health}. Our proposed need-to-know property follows the similar idea:
the system should use the minimum amount of information from users to provide a certain level of quality of service.
In a concurrent work Biega et al.~\cite{biega2020dataminimization} define the need to know principle
for computational applications and tie it to concepts in data protection laws.

\spara{Cost-sensitive learning and privacy as cost.}
Our definitions of fairness in privacy can be expressed in terms of a cost associated with each feature. 
This follows a line of research in machine learning that focuses on settings in which acquiring feature values is associated with some cost. The goal then is to make the best possible prediction with minimum cost users incurred at the test time. Some examples are decision trees with minimal cost~\cite{ling2004decision}, test-cost sensitive Naive Bayes classification~\cite{chai2004test},
and using a Markov decision process to sequentially acquire feature
values~\cite{shim2018joint,
maliah2018mdp, trapeznikov2013supervised}.

A number of previous papers have associated privacy more explicitly with a
cost~\cite{pattuk2015privacy, early2016test}.
Note however that while these works consider privacy as feature costs,
the general goal is that the privacy loss of each individual is minimized;
there is no consideration of fairness of privacy.

%A number of previous papers have associated privacy more explicitly with a cost.
%The authors in \cite{pattuk2015privacy} suggested considering
%privacy as a cost metric, which turns cost-sensitive methods into privacy preserving methods.
%%They propose a greedy approach to select a subset of features that maximize the  prediction confidence at the test time using a given privacy budget.
%Similarly, the authors in~\cite{early2016test} propose a framework for dynamically trading off feature cost against prediction quality.
%Note however that while the above works consider privacy as feature costs, the general goal is that the privacy loss of each individual is minimized;  there is no consideration of fairness of privacy.

\spara{Privacy and fairness.}
Recently both privacy and fairness researchers have recognized the importance
of understanding the interaction between privacy and fairness in algorithmic
decision systems~\cite{Cummings:2019:CPF:3314183.3323847,
jagielski2019differentially,bagdasaryan2019differential,hajian2015discrimination,
ruggieri2014anti}.
However, as eloquently argued by Ekstrand et. al.~\cite{ekstrand2018privacy},
much work remains to be done in ``characterizing under what circumstances
and definitions privacy and fairness are simultaneously achievable?''.
Our results in this paper can be seen as an effort to answer this question
by specifying some of the circumstances in which the interaction between
privacy and fairness can be formally studied.

The authors in~\cite{ekstrand2018privacy} also interpret fair privacy as
whether a privacy scheme protects all individuals equally, and they raise
questions about the implications of this property on other fairness notions;
however their discussion remains at a high level.

From a practical viewpoint, some studies have proposed techniques to
improve fairness and privacy at the same time~\cite{hajian2015discrimination,
hajian2012injecting, ruggieri2014anti}.
Furthermore, the authors in~\cite{pratesi2018prudence, pratesi2020primule}
provide a framework for empirical assessment of privacy risks associated with
different individuals when different subsets (dataviews) of a dataset are used.
This framework allows studying the trade-off between privacy risk and data utility
(which in turn is linked to accuracy). However, they do not consider an explicit notion
of fairness in privacy or in accuracy.

\spara{Differential Privacy and fairness.}
Another recent line of work considers a common pivacy notion, differential privacy~\cite{dwork2014algorithmic}, and studies its interaction with existing
fairness notions. In particular authors in~\cite{Cummings:2019:CPF:3314183.3323847}
prove that differential privacy is incompatible with satisfying equal false negative
rates among groups, and they provide a differentially private classification algorithm
that approximately satisfies group fairness guarantees with high probability.
Furthermore, it has been shown that applying differential privacy implies
unequal accuracy costs over different subgrups which results in decreasing fairness~\cite{bagdasaryan2019differential}.

In this paper, instead of differential privacy, we use a privacy notion
that is based on the set of revealed features of users,
which allows us to compare the privacy loss of different users.
%Dwork et al. in~\cite{dwork2012fairness} show that their notion of individual
%fairness can be interpreted as a generalization of differential privacy.
%However, this does not address how privacy and fairness interact.

%and study the relationship between the fairness in privacy and other properties of a decision system.

\spara{Incompatibility results.}
There are incompatibility results in the area of fairness in machine learning~\cite{chouldechova2017fair,kleinberg2016inherent}; however, they all consider different fairness measures defined for the output of a learning system, e.g., the trade-off between calibration, equal false positive rates, and equal false negative rates.  In contrast, this paper introduces a trade-off between fairness properties related to the outputs and inputs of a classifier.

%% file: Formulation-arXiv.tex
\section{Formulation and Setting}\label{sec:setting}
We start by establishing notation and a number of definitions.
We consider a set of features $F=\{f_1, \dots, f_d\}$ in which each feature $f_i$ takes values from the domain $\mathcal{F}_i$.
A dataset $\mathcal{D}$ is a set of data points (feature vectors) $\vecx_i \in \mathcal{X}$ where $\mathcal{X} = \mathcal{F}_1 \times \dots \times \mathcal{F}_d$ together with the corresponding labels $y_i \in \mathcal{Y}$, i.e., $\mathcal{D} \subset \{(\vecx_i,y_i)| \vecx_i \in \mathcal{X}, y_i \in \mathcal{Y}\}$.
For notational convenience, we use $\mathcal{D_X}$ to denote the set of feature vectors in $\mathcal{D}$, i.e.,
$\mathcal{D_X}=\{\vecx_i|\exists y \in \mathcal{Y} \ s.t. \  (\vecx_i,y) \in \mathcal{D}\}$. For any $S  \subseteq F$ and $\vecx_i$, $\Omega_S(\vecx_i)$ denotes feature vector $\vecx_i$ in which only values of the features in $S$ are revealed.

Let $X$ be a multivariate random variable that takes on values $\vecx \in \mathcal{D_X}$,
and $Y(X)$ be a random variable that denotes the true label of $X$ in $\mathcal{D}$. If no information about $X$ is known, the probability that the label of $X$ is $c \in \mathcal{Y}$ equals to\footnote{In this paper we assume a finite sample model using the given dataset. Thus our setting is an
instance of transductive learning as opposed to inductive learning in which the dataset is a sample from some distribution.}
\begin{equation*}
Pr[Y(X)=c] = \frac{|\{(\vecx,y) \in \mathcal{D} | y = c\}|}{|\mathcal{D}|}
\end{equation*}
This probability changes if some features values in $X$ are revealed. In particular, given that $\Omega_S(X)=\Omega_S(\vecx_i)$ we have:
\begin{equation*}
  \begin{split}
Pr [Y(X) = & c\, |\, \Omega_S(X)=\Omega_S(\vecx_i)] = \\
& \frac{|\{(\vecx,y) \in \mathcal{D} | \Omega_S(\vecx) = \Omega_S(\vecx_i) \wedge y = c \}|}{|\{(\vecx,y) \in \mathcal{D} | \Omega_S(\vecx) = \Omega_S(\vecx_i) \}|}
  \end{split}
\end{equation*}

A classifier $\hat{Y}$ is a function that predicts the label of a given
feature vector. We assume that $\hat{Y}$ is trained on all the features
in $F$, and at the test time it is applied to data points in a dataset $\mathcal{D}$.
Furthermore, We assume that $\hat{Y}$ can make a prediction using any
subset of the feature values (see footnote \ref{footnote1}). In particular,
$\hat{Y}(\Omega_S(\vecx_i))$ denotes the predicted label for $\vecx_i$
by $\hat{Y}$ using feature set $S$. We do not make any assumption about $\hat{Y}$ being a deterministic or a probabilistic function.

$\hat{Y}(X)$ is a random variable that denotes the label predicted for $X$ by $\hat{Y}$; similarly, $\hat{Y}(\Omega_S(X))$ is a random variable that denotes the label predicted for $X$ by the classifier $\hat{Y}$ based on the features in $S$.

\subsection{Predictive Power of a Feature Set}
For a given dataset $\mathcal{D}$, we define the \emph{predictive power} $\Phi_S(\vecx_i)$ of
a feature set $S \subseteq F$ for a data point $\vecx_i \in \mathcal{D_X}$, as the probability of the most probable label for $X$ given that the values of the features in $S$ are revealed
by $\vecx_i$, i.e., $\Omega_S(X)=\Omega_S(\vecx_i)$. In other words,
\begin{equation*}
\Phi_S(\vecx_i) = \max\limits_{c \in \mathcal{Y}} Pr[Y(X)=c |  \Omega_S(X) = \Omega_S(\vecx_i)]
\end{equation*}
If $\Phi_S(\vecx_i)=1$, we say that $\vecx_i$ is distinguishable in $\mathcal{D}$ using feature set $S$.

\subsection{Optimal Classifier}
We first define the accuracy of a classifier for a data point using a subset of features.

\spara{Prediction Accuracy.}
The \emph{accuracy} of the prediction $\hat{Y}(\Omega_S(\vecx_i))$ is the probability that the label predicted for $X$ using the features in $S$ is equal to the true label of $X$, given the feature values revealed by $\Omega_S(\vecx_i)$. In other words,
\begin{equation}\label{acc}
acc(\hat{Y}(\Omega_S(\vecx_i))) = Pr[\hat{Y}(\Omega_S(X)) = Y(X) |  \Omega_S(X) = \Omega_S(\vecx_i)].
\end{equation}
An optimal classifier is then defined as follows.

\spara{Optimal Classifier.}
Given a dataset $\mathcal{D}$, an optimal classifier $\hat{Y}_{opt}$ is a classifier that for all data points in $\mathcal{D}$ and using any subset of features $S \subseteq F$, has the highest prediction accuracy. In other words, $\hat{Y}_{opt}$ satisfies the following\footnote{This is an extension of the \emph{Bayes optimal classifier} to the settings where any subset of features can be used to make a prediction.}
\begin{equation*}
\forall \vecx_i \in \mathcal{D_X}, \forall S \subseteq F, \forall \hat{Y};\ acc(\hat{Y}(\Omega_S(\vecx_i)))
 \leq acc(\hat{Y}_{opt}(\Omega_S(\vecx_i)))
\end{equation*}
The following lemma provides a convenient way for computing the
accuracy of the predictions made by an optimal classifier.
In particular, it states that for any data point in a given dataset,
the accuracy of an optimal classifier using a set of features can be
computed by finding the predictive power of that feature set for the
corresponding data point. We later use this result to measure the performance of an optimal classifier by studying the characteristics of the dataset to which the classifier is applied.
\begin{lemma}\label{lemma}
A classifier is optimal for a given a dataset $\mathcal{D}$,
if and only if for any $\Omega_S(\vecx_i)$ it returns
the most probable label for $X$ given that $\Omega_S(X) = \Omega_S(\vecx_i)$.
\end{lemma}
\noindent The proof of Lemma \ref{lemma} is presented in appendix \ref{proof:lemma1}.

\begin{cor}\label{cor1}
The prediction accuracy of an optimal classifier for $\Omega_S(\vecx_i)$ is equal to the predictive power of set $S$ for $\vecx_i$,
i.e., $\Phi_S(\vecx_i)$.
\end{cor}

%% file: Properties.tex
\begin{table*}[t]
\centering
	\begin{minipage}[b]{.4\textwidth}
	\centering
	\caption{An illustrative dataset.}
	\label{table:dataset}
		\begin{tabular}{cccc}
		%\hline
		& \multicolumn{2}{c}{features}\\
		\cline{2-3}
		data point  & f1 & f2 & label \\
		\hline
		$\vecx_1$ & 0  & 0  & \color{black}{-}\\
		$\vecx_2$ & 0  & 1  & \color{black}{-}\\
		$\vecx_3$ & 0  & 3  & \color{red}{+} \\
		$\vecx_4$ & 2  & 3  & \color{black}{-} \\
		\hline
		\end{tabular}
	\end{minipage}
	\qquad
	\begin{minipage}[b]{.4\textwidth}
	\centering
	\caption{Predictive power of each feature set.}
	\label{table:predictive-power}
		\begin{tabular}{ccccc}
		& \multicolumn{4}{c}{feature sets}\\
		\cline{2-5}
		data point &
		$\Phi_\emptyset$ & $\Phi_{\{f1\}}$ & $\Phi_{\{f2\}}$ & $\Phi_{\{f1,f2\}}$\\
		\hline
		$\vecx_1$   & ${3}/{4}$ & ${2}/{3}$ & 1         & 1\\
		$\vecx_2$   & ${3}/{4}$ & ${2}/{3}$ & 1         & 1\\
		$\vecx_3$   & ${3}/{4}$ & ${2}/{3}$ & ${1}/{2}$ & 1\\
		$\vecx_4$   & ${3}/{4}$ & 1         & ${1}/{2}$ & 1\\
		\hline
		\end{tabular}
	\end{minipage}
	%\vspace{-0.5mm}% reduce too much white space after table 
\end{table*}
\raggedbottom % reduce too much white space after table
\section{Desired Properties}\label{sec:properties}
We now present formalizations of three properties that
involve privacy and fairness of classifiers.
The properties that we define in this section
depend on both the input features used, and the predictions made by a classifier.
For a dataset $\mathcal{D}$, we use $S_i \subset F$ to denote
the set of features used by a particular classifier to predict
the label of data point $\vecx_i \in \mathcal{D_X}$.
Note that the following properties can be validated for any arbitrary
choice of $S_i$ for each data point.

We emphasize that here we are not concerned about how $S_i$ is selected for a given
data point and a particular classifier, but we are rather concerned with
the social properties of using $S_i$ compared to other feature sets $S'_i \subset F.$
%In practice, for example, $S_i$ may be the result of a feature selection procedure.
(see footnote \ref{footnote2}.)
%such as those discussed in~\cite{maliah2018mdp,
%weiss2013learning, trapeznikov2013supervised}.

%\mc{Should optimality (3.2) be moved into this section?  It too is one
%  of the properties we are formalizing.}

\subsection{Output Property: Fair Prediction Accuracy}\label{sec:fair-accuracy}
In order to define a measure for the fairness in the outputs of a classifier,
we use the accuracy equality notion~\cite{verma2018fairness}, and extend
it to the individual level as has been suggested
in~\cite{speicher2018unified}. 

For a classifier $\hat{Y}$ and a dataset $\mathcal{D}$, let $S_i$ be the set of features used to
predict the label of $\vecx_i$. $\hat{Y}$ satisfies fair prediction accuracy if labels of all data points are predicted with equal accuracy, i.e.,
\begin{equation}\label{eq:fair-accuracy}
\exists \gamma \in (0,1] \ s.t. \ \forall \vecx_i \in \mathcal{D_X}, \  acc(\hat{Y}(\Omega_{S_i}(\vecx_i)))=\gamma
\end{equation}

\subsection{Input Property: Need to Know}
The need-to-know property states that for any data point,
using any proper subset of the features used by the classifier
will decrease the prediction accuracy (i.e., the feature set $S_i$
is minimal with respect to the prediction accuracy):
\begin{equation}\label{NK2}
\forall \vecx_i \in \mathcal{D_X}, \forall S' \subset S_i, \  acc(\hat{Y}(\Omega_{S'}(\vecx_i)))<acc(\hat{Y}(\Omega_{S_i}(\vecx_i))) 
\end{equation}

\if 0
Finally, the need-to-know property states that for any data point,
using any set of features whose cost is smaller than the cost of
features used by the classifier will decrease the prediction
accuracy. In other words, 
\begin{equation}\label{NK}
  \begin{split}
\forall & \vecx_i \in \mathcal{D_X}, \forall S \subseteq F, \\
&  \sum_{f_k \in S} \vecc(k) < \sum_{f_k \in S_i} \vecc(k) \Rightarrow acc(\hat{Y}(\Omega_{S}(\vecx_i)))<acc(\hat{Y}(\Omega_{S_i}(\vecx_i))) 
  \end{split}
\end{equation}

\noindent The specific types of cost vectors $\vecc$ lead to specific kinds of need-to-know:

\begin{description}
\item[Feature Count.] When $\vecc = c. \mathbf{1}_d,$ using any set of
  features whose size is smaller than the number of selected features by
  the classifier will decrease the prediction accuracy: 
\begin{equation}\label{NK1}
  \begin{split}
\forall & \vecx_i \in \mathcal{D_X}, \forall S \subset F, \\
&  |S| < |S_i| \Rightarrow acc(\hat{Y}(\Omega_{S}(\vecx_i)))<acc(\hat{Y}(\Omega_{S_i}(\vecx_i))) 
  \end{split}
\end{equation}

\item[Feature Match.] In this case, using any proper subset of the
  features selected by the classifier will decrease the prediction
  accuracy: 
\begin{equation}\label{NK2}
\forall \vecx_i \in \mathcal{D_X}, \forall S' \subset S_i, \  acc(\hat{Y}(\Omega_{S'}(\vecx_i)))<acc(\hat{Y}(\Omega_{S_i}(\vecx_i))) 
\end{equation}
\end{description}
\fi

Note that the need-to-know property does not imply that the prediction
accuracy must improve monotonically as the number of features that are used by the classifier increases.
Furthermore, although we consider accuracy as the criterion for which the use of data is minimized, other measures (e.g., false negative rate) may be more appropriate in specific applications. We leave studying the implications
of such alternative definitions of need-to-know for future work.

\subsection{Input Property: Fair Privacy}\label{sec:fair-privacy}
Fair privacy is determined by the input features used by classifier for each data point.
We assume each feature is associated with a non-negative cost that
denotes the privacy cost of revealing that feature, and that
the privacy cost of each feature is the same across all users. Let vector $\vecc \in \mathbb{R}_{\ge 0}^d$ denote the privacy costs of the features.
%We present examples of feature cost vectors below.

Fair privacy states that the total privacy costs of the used
features are equal for all data points, i.e., 
\begin{equation}\label{fair-privacy}
\exists \ell \in \mathbb{R} \ s.t. \ \forall \vecx_i \in \mathcal{D_X}, \  \sum\limits_{f_k \in S_i} \vecc(k) =\ell
\end{equation}

\noindent There are at least two natural cases for the cost vector $\vecc$:  

\begin{description}
\item[Feature Count.]  One may choose to treat the privacy costs of all features
  as equal.  Setting $\vecc = c. \mathbf{1}_d$ implies that privacy
  fairness holds when the number of used features is the same for all data points, i.e.,
\begin{equation}\label{fair-privacy1}
\exists k \in \mathbb{N} \ s.t. \ \forall \vecx_i \in \mathcal{D_X}, \  |S_i|=k
\end{equation}

\item[Feature Match.] Another natural approach is to treat two feature
  sets as equal-privacy-cost if and only if they contain the same features.
  This can be formalized by making the total cost of every subset of the
  features distinct (e.g. $\vecc = \{2^n| 0 \leq n \leq d-1\}$).   In
  this case, privacy fairness means that the exact same set of features
  are used to make a prediction for all data points:
\begin{equation}\label{fair-privacy2}
\exists S \subseteq F \ s.t. \ \forall \vecx_i \in \mathcal{D_X}, \  S_i = S
\end{equation}
\end{description} 

%% file: Trade-offs-arXiv.tex
\section{The Trade-off}\label{sec:trade-off}
In this section we study how the different socially important properties
of a classifier defined in Section \ref{sec:properties} interact.
In particular, since all the three properties have important social values,
it is natural to ask whether they can be satisfied simultaneously.
In other words, we ask whether it is possible for a classifier to use a
particular set of input features for each test instance, and satisfy all
three properties while maximizing prediction accuracy.

First, we show that there are situations (i.e., datasets) in which 
an optimal classifier cannot simultaneously satisfy
fair privacy, fair accuracy, and need-to-know.
Then we present a theorem that precisely characterizes all the datasets
in which such a trade-off exists under our definitions.  This implies
that in general, achieving fairness in the inputs and the outputs of an
optimal classifier are incompatible goals. 

\subsection{Presenting the Incompatibility}\label{sec:dataset}
\begin{figure}[h]
\centering
\includegraphics[width=0.3\textwidth]{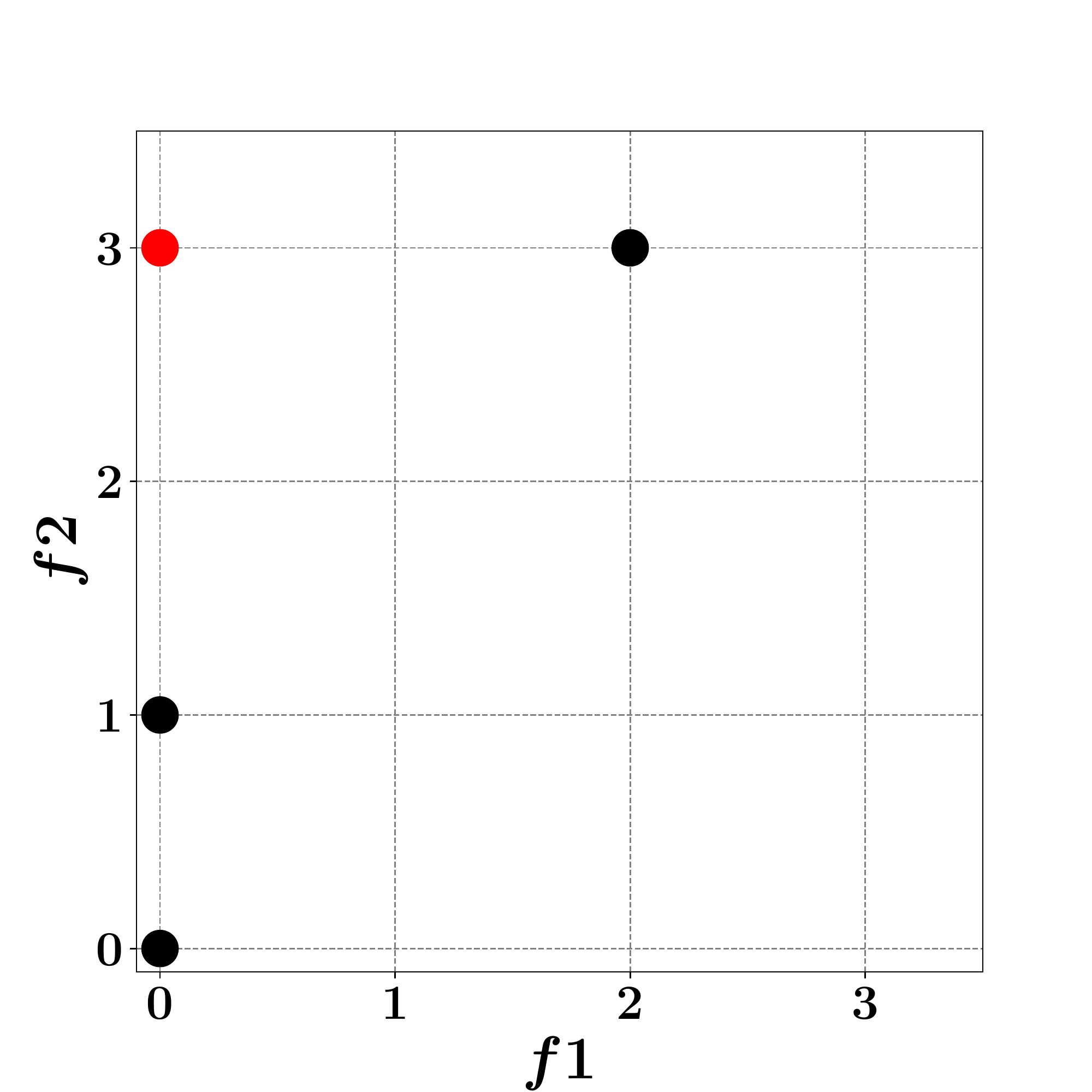}
\caption{}
\label{fig:dataset}
\end{figure}

We show that the following proposition is true:
\begin{prop}\label{prop1}
When applied to an arbitrary dataset, an optimal classifier cannot be
guaranteed to 
simultaneously satisfy fair privacy, fair accuracy, and need-to-know,
unless it is the trivial classifier that does not use any feature values
for all data points.
\end{prop}
%\begin{prop}\label{prop1}
%In general, an optimal classifier cannot simultaneously satisfy fair privacy,
%fair accuracy, and need-to-know when applied to an arbitrary dataset,
%unless it does not use any feature values from any data points.
%\end{prop}
\spara{Proof.}
We provide an example of a dataset for which any optimal classifier can satisfy
at most two of fair privacy, fair accuracy, and need-to-know.
Table \ref{table:dataset} presents our example dataset. The dataset contains
two features ($f_1$ and $f_2$), and four data points with class labels $y \in \{+,-\}$.
Figure \ref{fig:dataset} shows the data points in a 2D plane.
\raggedbottom

We present our arguments using two different feature cost vectors; each corresponds
to one of \emph{feature count} and \emph{feature match} cases introduced in section \ref{sec:fair-privacy}.
  
\textbf{Feature Count.} Assume that privacy costs of all features are 1, i.e., $\vecc = \mathbf{1}_2$.
Using corollary (\ref{cor1}), we know that the prediction accuracy of an optimal classifier for each data point
is equal to the predictive power of the selected feature set for that data point.
Table \ref{table:predictive-power} shows the predictive power of each subset of features for each
data point in the dataset.

First, assume an optimal classifier that satisfies fair privacy. Considering the given feature cost vector,
the privacy cost for each data point can be either 1 (the classifier uses either $f_1$ or $f_2$) or
2 (the  classifier uses both $f_1$ and $f_2$). (Notice that the case of using no feature values
for all data points is excluded from proposition\ref{prop1}.)
Therefore, in order to satisfy fair privacy, the privacy cost of all data points should be equal, and is either 1 or 2.

If the privacy cost is 1 for all data points (using either $f_1$ or $f_2$), from Table \ref{table:predictive-power}
we observe that it is not possible to have equal prediction accuracy for all data points. In particular,
the prediction accuracy for $\vecx_3$ is $\frac{2}{3}$ using $f_1$ and $\frac{1}{2}$ using $f_2$.
However, there is no way to have prediction accuracy of $\frac{2}{3}$ for $\vecx_4$, or $\frac{1}{2}$
for $\vecx_1$ and $\vecx_2$ using either $f_1$ or $f_2$. This violates fair prediction accuracy.

If the privacy cost is 2, all the labels can be predicted with accuracy $1.0$.
However, this violates the need to know property for data points $\vecx_1, \vecx_2, \vecx_4$
because the same prediction accuracy could be reached using only $f_2$ for
$\vecx_1$ and $\vecx_2$, or using $f_1$ for $\vecx_4$.

Therefore, any optimal classifier that satisfies fair privacy when applied on this dataset violates
either the fair prediction accuracy or the need to know property.  $\square$ 

\textbf{Feature Match.}
We could get the same result by assuming feature costs such that the total cost of every feature subset is distinct.
In that case, fair privacy reduces to using the same set of features for every data point. From Table \ref{table:predictive-power},
we observe that the only feature set that satisfies fair accuracy, i.e., the only column with equal predictive power for all data points,
is $\{f_1,f_2\}$; and using this set violates need-to-know for $\vecx_1, \vecx_2, \vecx_4$.

\subsection{Formal Specification}
The previous section presents a dataset for which at most two of the properties from Section~\ref{sec:properties} can be satisfied.
However, it remains to formalize when precisely a given dataset exhibits the trade-off, which we do in this section.
We do this for an optimal classifier under the Feature Match definition of fair privacy (eq.\ref{fair-privacy2}) and we
leave generalizing to other definitions of fair privacy for future work.
In the common case where privacy costs of the features are unknown, the
Feature Match definition--- i.e., using the same set of features for all
individuals--- is a reasonable choice.
Similar to Section \ref{sec:dataset}, in the following we exclude the
trivial classifier that does not use any feature values for all data points.

\begin{thm}\label{thm1}
There exists an optimal non-trivial classifier that satisfies fair privacy, fair accuracy, and need-to-know
when applied to a dataset $\mathcal{D}$,
if and only if $\mathcal{D}$ satisfies the following condition:
\begin{equation}\label{condition}
\begin{split}
\exists \ \text{``non-empty} \ S \text{''} \subseteq F \ & s.t.,\\
& \exists \gamma \in (0,1] \ s.t. \ \forall \vecx_i \in \mathcal{D_X}, \  \Phi_{S}(\vecx_i)=\gamma \\
& \wedge \\
& \forall \vecx_i \in \mathcal{D_X}, \forall S' \subset S, \ \Phi_{S'}(\vecx_i)<\Phi_{S}(\vecx_i)
\end{split}
\end{equation}
\end{thm}
\noindent The proof of Theorem \ref{thm1} is presented in appendix \ref{proof:theorem1}.

Theorem \ref{thm1} provides a necessary and sufficient condition (eq.\ref{condition}) to
identify datasets for which an optimal classifier can simultaneously satisfy all the three properties.
Notice that this condition is an statement about a dataset and can be verified independently
of any classifier. The statement can be written as the following description of a dataset:

\emph{``There is a non-empty feature set that has equal predictive power for all data points in the dataset. Furthermore, all subsets of that feature set have lower predictive power for all points in the dataset.''}

Consequently, the negation of (\ref{condition}) provides a necessary and sufficient condition
for the case where any optimal classifier can satisfy at most two of fair prediction accuracy, fair privacy, and need to know (i.e., datasets in which there is a trade-off between the abovementioned properties of any optimal classifier).
By negating (\ref{condition}) we find the following characterization of such datasets:
\begin{equation}\label{trade-off-condition}
\begin{split}
\forall \ \text{``non-empty} \ S \text{''} \subseteq & F,\\
& \exists \vecx_i, \vecx_j \in \mathcal{D_X} \ s.t. \ \Phi_{S}(\vecx_i) \neq \Phi_{S}(\vecx_j) \\
& \vee \\
& \exists \vecx_i \in \mathcal{D_X}, \exists S' \subset S \ s.t. \ \Phi_{S'}(\vecx_i) \geq \Phi_{S}(\vecx_i)
\end{split}
\end{equation}

For an intuitive interpretation of the above statement, assume an optimal classifier that satisfies fair privacy, i.e., set $S$ is used for all data points in the dataset. Therefore, in order to exhibit the trade-off, using $S$ the classifier should either violate fair prediction accuracy (first clause in (\ref{trade-off-condition})), or need-to-know (second clause in (\ref{trade-off-condition})). Thus, showing that for all non-empty $S \subseteq F$ either fair prediction accuracy or need-to-know are violated implies that no optimal
non-trivial classifier can satisfy all three properties.

\begin{cor}
Given a data set $\mathcal{D}$ and a non-trivial classifier $\hat{Y}$,
if $\mathcal{D}$ satisfies (\ref{trade-off-condition}) and $\hat{Y}$
satisfies fair privacy, fair accuracy, and need-to-know when applied to $\mathcal{D}$,
then $\hat{Y}$ is not optimal.
\end{cor}

%% file: Verification-arXiv.tex
\section{The Trade-off in Real Data}\label{sec:verification}
Given the results in the previous section, it is worthwhile to ask whether this trade-off is typical -- does it occur often in real-world data?   
We first develop a practical approach to answering this question for a given dataset, and then we apply our approach to various datasets from the standard UCI machine learning repository~\cite{Dua:2019}.
\begin{algorithm}[h]
 \small
 \KwIn{Dataset $\mathcal{D}$ with feature set $F$}
 \KwOut{Yes/No}
 \BlankLine
 initialize queue Q\\
 C = [ ]\\
 Q.put($\{\}$)\\
 \For{f in F}{
 \If{f has identical value over all data points}{
 remove f from F
 }
 
 }
 \While{Q is not empty}{
     S = Q.get()\\
     \If{$S \neq \emptyset$}{
     C.append(S)}
     compute $\Phi_{S}(\vecx_i)$ for all $\vecx_i \in \mathcal{D}$\\
     \If{$\Phi_{S}(\vecx_i) \neq 1$ for all $\vecx_i \in \mathcal{D}$}
         {\For{all features f in F whose index is larger than the largest index in S}
             {Q.put(S $\cup \{f\}$)\\
              }
         }
 }
 \For{candidate S in C}{
 \If{S satisfies the 1st clause in (\ref{trade-off-condition})}{
 continue
 }
 \If{S satisfies the 2nd clause in (\ref{trade-off-condition})}{
 continue
 }
 \Else{\KwRet{No}}
 }
 
 \KwRet{Yes}
 \BlankLine
 \caption{Verify if a given dataset holds the trade-off.}
\label{alg}
\end{algorithm}
\raggedbottom

\subsection{A Verification Algorithm}
For any given dataset, we may apply (\ref{trade-off-condition}) to test it, since  (\ref{trade-off-condition}) is a predicate that identifies all and only those datasets for which the trade-off is present.  
A naive approach to evaluating (\ref{trade-off-condition}) consists of computing $\Phi_{S_k}(\vecx_i)$ for all subsets $S_k \subseteq F$ and all $\vecx_i \in \mathcal{D}$. If for each $S_k$ at least one of the two clauses in (\ref{trade-off-condition}) are satisfied, no optimal classifier can simultaneously satisfy fair accuracy, fair privacy, and need-to-know when applied on $\mathcal{D}$.
However, the universal quantifier in (\ref{trade-off-condition}) implies a search over the exponential number of subsets in the power set of $F$.
Hence, we must consider how to efficiently verify that a given dataset satisfies
(\ref{trade-off-condition}). In this section, we introduce a verification algorithm that exploits several structures in the feature susbsets to prune the search space and is efficient in practice. 

The pseudocode of our dataset verification algorithm is provided in algorithm \ref{alg}.
The algorithm first generates feature subsets (candidates) for which an optimal classifier could possibly satisfy both fair accuracy and need-to-know. Then it eliminates each candidate that satisfies at least one of the two clauses in (\ref{trade-off-condition}). The algorithm uses an incremental method to generate candidates (i.e., larger sets are generated by adding more features to each of the existing candidates.) This allows the algorithm to recognize many of the candidates that will satisfy (\ref{trade-off-condition}) before actually generating them.
This is a key tool for pruning the search space and obtaining a practical algorithm.

The first pruning step is to notice that if a feature $f_i$ has identical values for all data points, removing $f_i$ from a feature subset $S$ does not change the predictive power of that feature subset.  That is, $\Phi_{S}(\vecx_i) =  \Phi_{S \setminus f_i}(\vecx_i)$ for all $\vecx_i \in \mathcal{D}$. Therefore, any feature subset that contains $f_i$ violates need-to-know. Consequently, we do not use such features in our candidate generation procedure (lines 4-6).  

The second pruning step is to notice that if $\Phi_{S_k}(\vecx_i)=1$ for some $S_k \subseteq F$ and some $\vecx_i$, then any superset $S^*_k$ of $S_k$ violates need to know property (i.e., second clause of (\ref{trade-off-condition})).  This is because predictive power cannot be larger than 1. Therefore, we can prune from our search space all the supersets of any feature set whose predictive power is 1 for at least one data point. As we generate new subsets, we compute the predictive power of each subset for all data points, and we stop adding more features to that subset once a data point is distinguishable in the dataset using that subset (lines 7-14). 

Finally, for each generated feature subset (candidate) we first verify the first
clause of (\ref{trade-off-condition});
if it is not satisfied we verify the second clause (lines 15-21).
Notice that the time complexity of verifying the first clause is linear in the size of the feature subset while the complexity of verifying the second clause is exponential in the size of the feature subset. If all the candidates satisfy at least one of the two clauses in (\ref{trade-off-condition}), we conclude that the given dataset holds the trade-off, i.e., an optimal classifier can satisfy at most two of fair accuracy, fair privacy, and need-to-know for the given dataset.

\subsection{Verifying Real Data}
Using Algorithm \ref{alg}, we find that it is possible to test reasonable-sized datasets and determine whether they exhibit the trade-off introduced in Section \ref{sec:trade-off}.
We obtain 18 datasets which have discrete feature domains from the UCI machine learning repository~\cite{Dua:2019}, and apply our verification algorithm to check if the trade-off
exists in each dataset.
Table \ref{table:UCI} in appendix \ref{results} summarizes
the datasets and the performance of the verification
algorithm for each dataset.

We observe that the size of the largest generated candidate for most of the datasets is significantly smaller than the number of features in that dataset, which shows that the superset pruning procedure is effective. The verification algorithm terminates in less than a minute for all cases even though a complete search over the power set of the features would be infeasible in most cases.  Also notice that except in one case (Nursery dataset), only verifying the first clause of (\ref{trade-off-condition}) is enough for all data points. 

Our algorithm verifies that every dataset we examined exhibits the trade-off between the three properties--- hinting
that the trade-off is prevalent in real-world data with discrete feature domains.
%The dataset contains 1797 data pints where each data point is an $8 \times 8$ image of a digit. Thus the dataset contains 64 features and each feature is an integer in range 0-16, and 10 class labels.
%
%while it is infeasible to iterate over all the $2^{64}$ subsets of the features, our incremental candidate generation algorithm terminates after generating 13210 feature subsets. The largest candidate set generated by our algorithm is:
%\texttt{['f1', 'f9', 'f25', 'f31', 'f33', 'f40', 'f64']}
%
%After removing all the candidates that satisfy the first clause of (\ref{trade-off-condition}), i.e., the predictive power of that feature set is different for at least two data points, the following feature subsets remain in the candidates:\\
%\texttt{['f1']}\\
%\texttt{['f33']}\\
%\texttt{['f40']}\\
%\texttt{['f1', 'f33']}\\
%\texttt{['f1', 'f40']}\\
%\texttt{['f33', 'f40']}\\
%\texttt{['f1', 'f33', 'f40']}\\
%
%Looking at the dataset, we realize that values of the feature \texttt{'f1'}, \texttt{'f33'}, and \texttt{'f40'} are 0 for all data points in the dataset. Thus, the predictive power of all the candidates that only contain these features is equal to the predictive power of the empty set. Therefore all the above candidates violate need to know property (i.e., satisfy the second clause in (\ref{trade-off-condition})) and the hand written digits data set holds the trade off introduced in section 5.

%% file: Conclusion-arXiv.tex
\section{Discussion and Concluding Remarks}\label{sec:conclusion}
In this paper we argue that the fairness notions for
algorithmic decision-making systems should expand to
incorporate the inputs (i.e., features) used by a system,
and we formulate two of such input properties: \emph{fair privacy} and \emph{need-to-know}.

We prove that in general an \emph{optimal} non-trivial classifier
cannot satisfy all of fair privacy, need-to-know, and fair accuracy.
Furthermore, we characterize all the datasets in which
the above trade-off exists using logical predicates.
Finally, we provide an algorithm that exploits several
computational efficiencies to verify if the trade-off
is present in a given dataset.
\begin{figure}[h]
%\begin{wrapfigure}{R}{0.34\textwidth}
\centering
\includegraphics[width=0.35\textwidth]{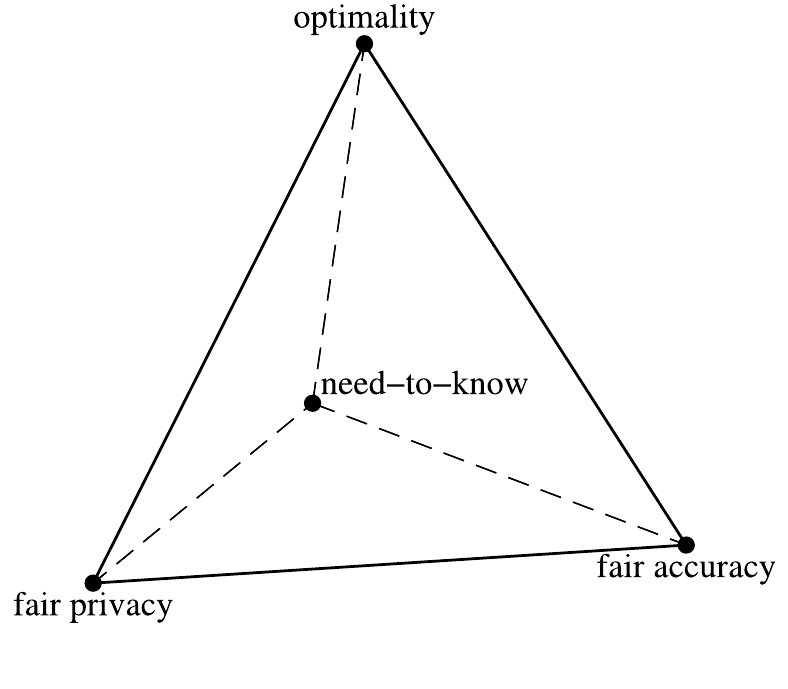}
%\caption{Tetrahedron in which any three properties may be achievable.}
\caption{}
\label{fig:tetrahedron}
%\vspace*{-15pt}
\end{figure}
%\end{wrapfigure} 
\raggedbottom

The tetrahedron in Figure \ref{fig:tetrahedron} can be used to
summarize our results.
In particular, in general the properties at all four vertices
cannot be satisfied together.
Moreover, each vertex offers a potentially interesting
direction for future exploration.

First, if one sets aside optimality to achieve the three socially desirable properties,
the question arises then how close to optimal can the performance of such a fair-input and and fair-output classifier be on a given
dataset\footnote{An example of such non-optimal classifier is provided in appendix \ref{sec:nonoptimal}.}.

Second, if one instead sets aside fair privacy, one may seek to achieve the other goals, perhaps following in the general style taken in~\cite{noriega2019active}, i.e., using different input features from different individuals.

Third, one may rather choose to set aside need-to-know.
For example, the authors in~\cite{canetti2019soft} equalize false positive, false negative, false discovery, and false omission rates
across the protected groups by deferring on some decisions (i.e., avoid making a decision for some individuals).
However, deferred decisions violate our need-to-know principle
which requires the system to use only data inputs that are
necessary for improving its predictions.

Finally, one may set aside fair accuracy, perhaps in favor of weaker conditions
such as fair mistreatment~\cite{zafar2017fairnessb,canetti2019soft}.
In that case, the question remains open whether other properties are achievable.

%% file: acks.tex
%\medskip \noindent \textbf{Acknowledgements .}
\begin{acks}
This research was supported by ERC Advanced Grant ``Foundations for Fair Social Computing'' (No. 789373), and the National Science Foundation under grant numbers IIS-1421759 and CNS-1618207.
\end{acks}

%% file: proofs.tex
\section{Proofs}

\subsection{Lemma 1}\label{proof:lemma1}
We write the right hand side of (\ref{acc}) as:\begin{equation}\label{acc2}
\sum_{c \in \mathcal{Y}} Pr[\hat{Y}(\Omega_S(X))=c, Y(X)=c  |  \Omega_S(X) = \Omega_S(\vecx_i)]
\end{equation}
Since $\hat{Y}(\Omega_S(X))$ only depends on the values
of the features in $S$, $\hat{Y}(\Omega_S(X))$ and $Y(X)$ are conditionally independent given a set of fixed values $\Omega_S(X) = \Omega_S(\vecx_i)$. Therefore (\ref{acc2}) is equal to:
\begin{equation}\label{acc-expand}
\sum_{c \in \mathcal{Y}} Pr[\hat{Y}(\Omega_S(X))=c | \Omega_S(X) = \Omega_S(\vecx_i)] Pr[Y(X)=c | \Omega_S(X) = \Omega_S(\vecx_i)]
\end{equation}

For any $\Omega_S(\vecx_i)$, let $p_c=Pr[Y(X)=c |  \Omega_S(X) = \Omega_S(\vecx_i)] $ and $p^* = \max\limits_{c \in \mathcal{Y}} p_c$. Also let $\hat{p}_c = Pr[\hat{Y}(\Omega_S(X))=c | \Omega_S(X) = \Omega_S(\vecx_i)]$ for an arbitrary classifier $\hat{Y}$. Using (\ref{acc-expand}) we can write the following for any classifier $\hat{Y}$:

\begin{equation*}
acc(\hat{Y}(\Omega_S(\vecx_i))) = \sum_{c \in \mathcal{Y}} p_c \hat{p}_c \leq \sum_{c \in \mathcal{Y}} p^* \hat{p}_c = p^*
\end{equation*}
Therefore, $p^*$ is an upper bound for the prediction accuracy of any classifier applied on $\Omega_S(\vecx_i)$.
Now let $c^* = \arg\max\limits_{c \in \mathcal{Y}} p_c$, the prediction accuracy of a classifier that deterministically returns $c^*$ for $\Omega_S(\vecx_i)$ (i.e., $\hat{p}_c =0$ for all $c \neq c^*$, and $\hat{p}_{c^*} = 1$) is $p^*$; therefore, such classifier is optimal.

Finally, we show that the prediction accuracy of any classifier with $\hat{p}_{c^*} < 1$ is lower than $p^*$. Assume a classifier $\hat{Y}'$ for which $\hat{p}_{c^*} = 1 - \epsilon$ and $\hat{p}_{c'} = \epsilon$ for some $\epsilon>0$ and some $c' \in \mathcal{Y}$ such that $p_{c'} < p^*$. Thus,
\begin{equation*}
acc(\hat{Y}'(\Omega_S(\vecx_i))) = p^* (1 - \epsilon) + p_{c'} \epsilon = p^* + \epsilon (p_{c'} - p^*) < p^* \  \ \square
\end{equation*}

\subsection{Theorem 1}\label{proof:theorem1}
For a classifier $\hat{Y}$ applied to a dataset $\mathcal{D}$, let $S_i$ denote the
set of features used from $\vecx_i$. First we repeat and name the following definitions
from section \ref{sec:properties},

\begin{description}
\setlength\itemsep{3pt}
\item[Fair Accuracy] \emph{(p1)}: \\
$\exists \gamma \in (0,1] \ s.t. \ \forall \vecx_i \in \mathcal{D_X}, \ acc(\hat{Y}(\Omega_{S_i}(\vecx_i)))=\gamma$
\item[Fair Privacy]  \emph{(p2)}: \\
$\exists S \subseteq F \ s.t. \ \forall \vecx_i \in \mathcal{D_X}, \ S_i = S$
\item[Need-To-Know]  \emph{(p3)}: \\
$\forall \vecx_i \in \mathcal{D_X}, \forall S' \subset S_i, \ acc(\hat{Y}(\Omega_{S'}(\vecx_i)))<acc(\hat{Y}(\Omega_{S_i}(\vecx_i))) $
\end{description}

Let $\mathcal{H}_{opt}$ be the set of all optimal non-trivial classifiers for $\mathcal{D}$. Assume there exists an optimal non-trivial classifier that satisfies all of \emph{p1},  \emph{p2}, and \emph{p3} when applied on $\mathcal{D}$, i.e., the following statement is true:
\begin{equation}\label{all3}
\exists \hat{Y} \in \mathcal{H}_{opt} \ s.t. \ p1 \wedge p2 \wedge p3
\end{equation}
From \emph{p2} we infer that the same set of features is used by the classifier to predict the label of all the data points. We call this set $S$ and we replace $S_i$ with $S$ in \emph{p1} and \emph{p3} ($S$ is non-empty since $\hat{Y}$ is non-trivial). Moreover, since $\hat{Y}$ is an optimal classifier, by Corollary (\ref{cor1}) we can replace the prediction accuracy of $\hat{Y}$ for any data point and any feature set with the predictive power of that feature set for that data point. Thus, from (\ref{all3}) we infer that the following statement is true:
\begin{equation}\label{necessary}
\begin{split}
\exists  \ \text{``non-empty} \ S \text{''} \subseteq F & \  s.t.,\\
&\exists \gamma \in (0,1] \ s.t. \ \forall \vecx_i \in \mathcal{D_X}, \  \Phi_{S}(\vecx_i)=\gamma\\
&\wedge\\ 
&\forall \vecx_i \in \mathcal{D_X}, \forall S' \subset S, \ \Phi_{S'}(\vecx_i)<\Phi_{S}(\vecx_i)
\end{split}
\end{equation}

On the other hand, assume we are given a dataset for which statement (\ref{necessary}) is true. We can then define a classifier $\hat{Y}$ such that it
uses the features in $S$ for all $\vecx_i \in \mathcal{D}$, and it returns $\arg\max\limits_{c \in \mathcal{Y}} Pr[Y(X)=c |  \Omega_{S_k}(X) = \Omega_{S_k}(\vecx_i)]$ for any $S_k \subseteq F$ and $\vecx_i \in \mathcal{D}$, i.e., $\hat{Y}$ is optimal. Therefore, $acc(\hat{Y}(\Omega_{S_k}(\vecx_i))) = \Phi_{S_k}(\vecx_i)$ and from (\ref{necessary}) we infer that $\hat{Y}$ satisfies $p1$ and $p3$. Furthermore, $\hat{Y}$ satisfies $p2$ because it uses $S$ for all data points, and is optimal by definition. Therefore, statement (\ref{all3}) is satisfied for $\mathcal{D}$.

Thus the property defined in (\ref{necessary}) is a necessary and sufficient condition to recognize datasets for which there exists an optimal classifier that satisfies all properties \emph{p1}, \emph{p2}, and \emph{p3}. $\square$

%% file: nonoptimal.tex
\section{A non-optimal classifier that satisfies all the three properties}\label{sec:nonoptimal}
As we discussed in Section \ref{sec:conclusion}, one may give up optimality in
order to achieve a classifier that simultaneously satisfies all the three properties
defined in Section \ref{sec:properties}.
In this section, we present an example of such a non-optimal classifier for the
dataset in Table \ref{table:dataset}.

We use a probabilistic classifier for our discussion.
Let $\vecx.f_i$ denote the value of feature $f_i$ in data point $\vecx$.
Now consider the classifier defined by equation (\ref{nonoptimal-def}).
This classifier first selects a linear classifier based on the set of known
feature values $S$. Then it returns a binary label using an additional randomization step.  
\begin{equation}\label{nonoptimal-def}
\hat{Y}(\Omega_S(\vecx)) =
  \begin{cases}
    S=\{f_1,f_2\} \quad \begin{cases} 
                         \vecx.f_2 - \vecx.f_1 \geq 2   &\quad +   \quad w.p. \frac{4}{5} \\
       					  otherwise					      &\quad - \quad w.p. \frac{4}{5} \\				
        				 \end{cases}\\
        				\\
    S=\{f_1\}     \qquad \begin{cases}
                         \vecx.f_1 \geq 1    &\quad - \quad w.p. \frac{3}{4} \\
       					  otherwise			  &\quad +	\quad w.p. \frac{3}{4} \\     
        				 \end{cases}\\
        										\\
    S=\{f_2\}     \qquad \begin{cases}
                         \vecx.f_1 \geq 2    &\quad + \quad w.p. \frac{3}{4} \\
       					  otherwise			  &\quad -	\quad w.p. \frac{3}{4} \\     
        				 \end{cases}\\
        				 \\
    S=\emptyset   \qquad  \begin{cases} 
       											     + \quad w.p. \frac{1}{2} \\
       											     - \quad w.p. \frac{1}{2} \\
        										\end{cases}
  \end{cases}
\end{equation}

Table \ref{table:non-optimal} shows the accuracies of the predictions made by this classifier
for each data point and each feature set. The values in the table are  calculated using equation (\ref{acc}).
First observe that by using feature set $\{f_1,f_2\}$ for all the data points, the classifier satisfies
fair privacy.
Furthermore, the accuracy of the classifier for all data points using this feature set is $\frac{4}{5}$,
meaning that fair accuracy is also satisfied.
Finally, we observe that using any subset of $\{f_1,f_2\}$ will result in  a lower prediction accuracy
for all the data points in the dataset, thus the need-to-know principle is also satisfied.

In order to see that the classifier defined by equation (\ref{nonoptimal-def}) is not optimal,
notice that our optimality definition requires the classifier to be optimal for all data points
and using any subset $S \subset F$.
However, using $\{f_1,f_2\}$ we see that the accuracy of the classifier for all
the data points is $\frac{4}{5}$ while all the data points are distinguishable using $\{f_1,f_2\}$,
i.e., the accuracy of an optimal classifier using $\{f_1,f_2\}$ is 1 for all data points in the dataset.
\begin{table}[t]
\caption{Accuracies of the predictions made by the classifier in equation (\ref{nonoptimal-def}).}
\label{table:non-optimal}
\centering
\begin{tabular}{c|c|c|c|c}
   & $\emptyset$ & ${\{f1\}}$ & ${\{f2\}}$ & ${\{f1,f2\}}$ \\
\hline
$\vecx_1$        & ${1}/{2}$ & ${5}/{12}$ & ${3}/{4}$ & ${4}/{5}$\\
$\vecx_2$        & ${1}/{2}$ & ${5}/{12}$ & ${3}/{4}$ & ${4}/{5}$\\
$\vecx_3$        & ${1}/{2}$ & ${5}/{12}$ & ${1}/{2}$ & ${4}/{5}$\\
$\vecx_4$        & ${1}/{2}$ & ${3}/{4}$  & ${1}/{2}$ & ${4}/{5}$\\
\end{tabular}
\end{table}

This example illustrates an interesting direction for future research, which can be
stated as the following question:
\emph{``how much one needs to compromise on optimality in order to simultaneously
achieve all the three socially desirable properties introduced in this paper?''}
%%\begin{wrapfigure}{R}{0.4\textwidth}
%\begin{figure}[h]
%\centering
%\includegraphics[width=0.28\textwidth]{figures/cartesian}
%\caption{}
%\label{fig:dataset}
%%\end{wrapfigure}
%\end{figure}

%% file: results.tex
\section{Dataset Verfification Resutls}\label{results}
\begin{table}[h]
\caption{The results of applying our verification algorithm to 18 UCI datasets.}
\label{table:UCI}
\centering
\resizebox{0.95\textwidth}{!}{
\begin{tabular}{l|c|c|c|c|c|c|c}

Dataset & size & \# features & \# labels &  \shortstack{largest\\ candidate size} & \shortstack{all candidates satisfy\\ the 1st condition}  & \shortstack{both 1st and 2nd\\ conditions were verified} & trade-off\\
\hline
Handwritten Digits & 1797 & 64 & 10 & 4 &     \Checkmark & \ding{56} & YES\\
Haberman's Survival & 306 & 3 & 2 & 2 &   	   \Checkmark & \ding{56} & YES\\
Letter Recognition & 20000 & 16 & 26 & 2 &    \Checkmark & \ding{56}& YES\\
Somerville Happiness & 143& 6 & 2 & 2 & 		\Checkmark & \ding{56} & YES\\
Vehicle Silhouettes & 846 & 18 & 4 & 2 &		\Checkmark & \ding{56} & YES\\
Caesarian & 80 & 5 & 2 & 3 &					\Checkmark & \ding{56} & YES\\
Musk (Version 1) & 476 & 166 & 2 & 1 &			\Checkmark & \ding{56} & YES\\
Musk (Version 2) & 6598 & 166 & 2 & 1 &		\Checkmark & \ding{56} & YES\\
Optical Digits & 3823 & 64 & 10 & 2 &			\Checkmark & \ding{56} & YES\\
Pen-Based Digits & 7494 & 16 & 10 & 2  &		\Checkmark & \ding{56} & YES\\
Mushroom & 5644 & 22 & 2 & 3 &					\Checkmark & \ding{56} & YES\\
Nursery & 12960 & 8 & 5 & 8 &		 \ding{56}  & \Checkmark & YES\\
Census Income & 32561 & 14 & 2 & 3 &			\Checkmark & \ding{56} & YES\\
Chess & 28056 & 6 & 18 & 5 &					\Checkmark & \ding{56} & YES\\
Contraceptive Method & 1473 & 9 & 3 & 4 &      \Checkmark & \ding{56} & YES\\
Balance Scale & 625 & 4 & 3 & 3 &              \Checkmark & \ding{56} & YES\\
Breast Cancer & 277 & 9 & 2 & 4 &              \Checkmark & \ding{56} & YES\\
Car Evaluation& 1728 & 6 & 4 & 4 &             \Checkmark & \ding{56} & YES\\
\end{tabular}
}
\end{table}